\documentclass[acmsmall]{acmart}
\usepackage{makecell}
\usepackage{multirow}
\usepackage{array} 
\usepackage{graphicx}
\AtBeginDocument{%
  \providecommand\BibTeX{{%
    \normalfont B\kern-0.5em{\scshape i\kern-0.25em b}\kern-0.8em\TeX}}}

\setcopyright{acmlicensed}
\copyrightyear{2023}
\acmYear{2023}
\acmDOI{XXXXXXX.XXXXXXX}

\begin{document}

%%
%% The "title" command has an optional parameter,
%% allowing the author to define a "short title" to be used in page headers.

\title{Why Do Large Language Models (LLMs) Struggle to Count Letters?}

%%
%% The "author" command and its associated commands are used to define
%% the authors and their affiliations.
%% Of note is the shared affiliation of the first two authors, and the
%% "authornote" and "authornotemark" commands
%% used to denote shared contribution to the research.

\author{Tairan Fu}
\email{ftr258@nuaa.edu.cn}
\affiliation{
  \institution{College of Mechanical \& Electrical Engineering, Nanjing University of Aeronautics and Astronautics}
  \streetaddress{Yudao Street 29}
  \city{Nanjing}
  \state{Jiangsu}
  \country{China}
  \postcode{210001}
}

\author{Raquel Ferrando}
\email{raquel.ferrando@alumnos.upm.es}
\author{Javier Conde}
\email{javier.conde.diaz@upm.es}
\author{Carlos Arriaga}
\email{carlos.arriaga.prieto@upm.es}
\author{Pedro Reviriego}
\email{pedro.reviriego@upm.es}
\affiliation{
  \institution{ETSI de Telecomunicaci\'on, Universidad Polit\'ecnica de Madrid}
  \streetaddress{Avda Complutense 30}
  \city{Madrid}
  \state{Madrid}
  \country{Spain}
  \postcode{28040}
}

%%
%% By default, the full list of authors will be used in the page
%% headers. Often, this list is too long, and will overlap
%% other information printed in the page headers. This command allows
%% the author to define a more concise list
%% of authors' names for this purpose.
%\renewcommand{\shortauthors}{Trovato and Tobin, et al.}

%%
%% The abstract is a short summary of the work to be presented in the
%% article.
\begin{abstract}

Large Language Models (LLMs) have achieved unprecedented performance on many complex tasks, being able, for example, to answer questions on almost any topic. However, they struggle with other simple tasks, such as counting the occurrences of letters in a word, as illustrated by the inability of many LLMs to count the number of "r" letters in "strawberry". Several works have studied this problem and linked it to the tokenization used by LLMs, to the intrinsic limitations of the attention mechanism, or to the lack of character-level training data. In this paper, we conduct an experimental study to evaluate the relations between the LLM errors when counting letters with 1) the frequency of the word and its components in the training dataset and 2) the complexity of the counting operation. We present a comprehensive analysis of the errors of LLMs when counting letter occurrences by evaluating a representative group of models over a large number of words. The results show a number of consistent trends in the models evaluated: 1) models are capable of recognizing the letters but not counting them; 2) the frequency of the word and tokens in the word does not have a significant impact on the LLM errors; 3) there is a positive correlation of letter frequency with errors, more frequent letters tend to have more counting errors, 4) the errors show a strong correlation with the number of letters or tokens in a word and 5) the strongest correlation occurs with the number of letters with counts larger than one, with most models being unable to correctly count words in which letters appear more than twice. These results suggest that the problems of LLMs to count letters are not related to the frequency of words or tokens in the training data but to the complexity of the counting operation. However, further studies are needed to build a better understanding of the limitations of LLMs to count the letters in a word.   
  
\end{abstract}

%%
%% The code below is generated by the tool at http://dl.acm.org/ccs.cfm.
%% Please copy and paste the code instead of the example below.
%%



%%
%% Keywords. The author(s) should pick words that accurately describe
%% the work being presented. Separate the keywords with commas.

%\received{20 February 2007}
%\received[revised]{12 March 2009}
%\received[accepted]{5 June 2009}

%%
%% This command processes the author and affiliation and title
%% information and builds the first part of the formatted document.
\maketitle

\section{Introduction}

Since the introduction of ChatGPT, the adoption of large language models (LLMs) has been exponential and are used by millions of people \cite{ChatGPT_survey}. This is because LLMs have achieved an unprecedented performance on many language processing tasks, conversational chatbots being just one of them. LLMs can, for example, answer multiple choice questions on almost any topic, and more difficult tests have to be designed as LLMs approach 100\% accuracy on existing tests \cite{mmlu_pro}. LLMs have also shown good results in solving mathematical problems \cite{Mathmeasuring} and also in solving common sense questions \cite{CommonSensemeasuring}. Finally, many LLMs are capable of performing these tasks not only in English, they support many languages although with a performance loss. Despite their good results in tasks that seem complex to humans, LLMs struggle with tasks that appear to be much easier, such as computing the result of simple arithmetic operations \cite{LLMmath} or counting the letters in a word \cite{CountLetters1}. In a way, LLMs behave like Ireneo Funes, the protagonist of "Funes the memorious" a short story by Jorge Luis Borges \cite{borges1962funes}. Irineo had an almost infinite memory capable of remembering every instant of his existence with every detail and this prevented him from reasoning \cite{funes_analysis} as Borges explains "Without effort, he had learned English, French, Portuguese, Latin. I suspect, nevertheless, that he was not very capable of thought. To think is to forget a difference, to generalize, to abstract. In the overly replete world of Funes there were nothing but details, almost contiguous details". LLMs know the details of almost anything we can ask on different languages but yet like Irineo struggle to count the letters in Strawberry.   

This performance in complex tasks is achieved by training large models with billions of parameters with datasets of trillions of words \cite{llm_general_survey}. The LLM learns to predict the next element in the sentence, and this process is repeated to generate the requested text or the answer to a given question. Most LLMs are currently based on the same architecture, the transformer \cite{vaswani2023attention} that is composed of layers, each of which implements a neural network and a new mechanism called attention to better capture the relationships between parts of the text. The size of the model can be scaled by adding more layers or making each of them more complex, resulting in models that exceed a trillion parameters. 

It would seem natural to use words as elements of a sentence when the LLM makes predictions. However, this is not efficient as it would require the LLM to estimate the probabilities for all words to make the decision. The number of words is very large, especially when you consider the variations of words such as verbs and the number of languages supported by LLMs. On the other extreme, LLMs could use letters as the elements to predict, thus radically reducing the number of possible outcomes. However, that would also be inefficient, as now many predictions will be needed to generate text, each requiring a significant computational effort. Interestingly, similar trade-offs apply also to human writing systems. Learning the letters of an alphabet, typically tens of letters, is much easier than learning the characters that describe concepts in languages such as Chinese with thousands of characters. Other languages use syllabaries or ideograms for writing having an intermediate number of basic elements or graphemes. However, once learned, a larger number of graphemes can be more efficient in processing the information although the optimality of the writing system has many different angles \cite{baroni2011alphabetic}.      

Most LLMs adopt a compromise solution and use sequences of characters, known as tokens, that are part of a word or an entire word \cite{Tokenizer1} for predictions. The tokenization tries to reduce the number of predictions needed to complete a text by assigning tokens to sequences of characters that appear frequently in the text. For example, ``strawberry'' is decomposed by the GPT4-o tokenizer into three tokens: ``st'', ``raw'' and ``berry''. The design of a tokenizer is a complex problem and many algorithms have been proposed in the literature \cite{TokenizerTheo}. One concern for tokenization is also to provide comparable performance in different languages \cite{TokenizerLanguages}. The design of the tokenizers affects the quality of responses in different languages and also the response time, cost, and energy consumption. Some tokenizers favor certain languages, which means that the same text in one language may require more tokens (inferences) than in another for its generation \cite{NEURIPS2023_74bb24dc}. For example, the sentence ``The sun is brilliant'' which contains 20 characters including spaces, is composed of 4 tokens according to the GPT-3 tokenizer (``The'', ``sun'', ``is'', ``brilliant''). However, its Spanish version, ``El sol es brillante'' contains 19 characters and is divided into 5 tokens (``El'', ``sol'', ``es'', ``brill'', ``ante'').

The fact that LLMs work with tokens means that as opposed to humans, letters are not their basic element for constructing words and then text. For us, learning letters and how to form words and pronounce them is a fundamental part of language acquisition \cite{barratt2020literacy}. Therefore, we identify letters from an early age and find it trivial to count the number of letters in a word. This is not the case for LLMs and it has been speculated that it can be due to the use of tokens instead of letters \cite{CountLetters3}. To the best of our knowledge, no comprehensive study on the errors made by LLMs has been reported in the literature. Such a study could shed light on the underlying factors that prevent LLMs from correctly counting the letters in a word. This paper evaluates a representative group of LLMs when counting letters on a large number of words and explores the relationships of those errors with different word parameters such as the number of letters or the word frequency. 

The remainder of the paper is organized as follows. Section \ref{sec:RelatedWord} covers related works on the limitations of LLMs for simple tasks such as counting letters. The methodology used in our evaluation is described in Section \ref{sec:EvalMethod} covering the LLMs used, the word features considered, the words evaluated, and the experimental procedure. The results are presented, analyzed and discussed in section \ref{sec:Results}. The paper ends with the conclusion in Section \ref{sec:Conclusion}.

\section{Related work}
\label{sec:RelatedWord}

The limitations of LLMs to perform some simple tasks are well-known. Those include simple arithmetic operations \cite{LLMmath} and counting of letters in a word or text \cite{CountLetters1},\cite{CountLetters2}. This issue was brought to the spotlight with the inability of leading LLMs to count the number of ``r'' letters in ``strawberry'', a task that is easily done by a child. The limitation not only applies to counting letters but also to other operations at the character level such as spelling or manipulation \cite{edman-etal-2024-cute}.

The problems of LLMs to count letters have been analyzed theoretically in \cite{CountLetters2} which shows that transformer-based LLMs are constrained in the number of letters that they can count by the size of certain model parameters related to the attention mechanisms and the model embeddings. However, many current models have sizes of those parameters that are above those values. Other works link the inability of LLMs to count letters to the use of tokens instead of characters as the processing units, arguing that LLMs in most cases do not work with letters, only when the token happens to be a letter \cite{CountLetters3}. This means that LLMs see few letter-level data during their training which can also limit their ability to count letters. However, there are also recent studies that challenge those conjectures \cite{CountLetters1} and propose the use of ``reasoning before answering'' to have correct results when asking LLMs to perform operations on letters. 

The analysis of previous works show that there is still a lack of understanding of why LLMs struggle to count letters. In this work, we contribute to the study of this problem by exploring the relation of the LLM errors with different word features, as discussed in the following sections.

\section{Methodology}
\label{sec:EvalMethod}

This section discusses the methodology used in the evaluation, including the selection of LLMs and words to be used, as well as the procedures used to gather and process the data and the features analyzed. Intuitively, the features that could affect the ability of LLMs to count the letters in a word are: 1) those related to the frequency of the word, tokens, and letters on the training dataset of the LLM and 2) those related to the difficulty of counting letters in a word, such as the number of letters, tokens, number of distinct letters, and letter counts.

\subsection{Procedure}

To automate the evaluation process, we ask the LLMs to count the number of letters that appear on each word and produce a JSON that contains the letter and their counts. The prompt used is: `Count the frequency of each letter in the input word and output it in JSON format, generate a JSON format reply directly without any additional information required. Here's an example:
        Input: strawberry
        Output: \{"s": 1, "t": 1, "r": 3, "a": 1, "w": 1, "b": 1, "e": 1, "y": 1\}'

This JSON is then compared with the correct counts of the letters for that word and if there is any difference, the word is reported as erroneously counted. The JSONs of all the responses are saved so that further analysis, for example, of the errors per letter or per count value, can be done.

\subsection{Frequency features}

The frequency of a word and its components in the training dataset can impact the ability of the LLM to learn words. For example, LLMs tend to achieve better performance in English, which typically accounts for the majority of the training text \cite{openai2023gpt4}. To see if this is the case when counting letters, we explore the relation of counting errors with frequency of words, tokens, and letters. Unfortunately, in most cases, the training datasets used in the LLMs are not available but we know that the models have been trained on massive amounts of data taken from the Internet. To overcome this limitation in accessing training data, we use the results of the analysis of a one trillion words dataset that is publicly available\footnote{\url{https://research.google/blog/all-our-n-gram-are-belong-to-you/}} which provides word and letter frequencies \cite{norvig2009natural}. For each LLM we use its tokenizer to decompose each word in tokens and then, based on that list, we also compute the relative token frequency.  

% The data is here: https://norvig.com/ngrams/

In summary, we use the following frequency-related features:

\begin{enumerate}
    \item Word frequency
    \item Token frequency
    \item Letter frequency
\end{enumerate}

\subsection{Count difficulty features}

The difficulty of counting the letters in a word can be related to the number of letters in the word but also to the number of letters that appear more than once. For an LLM since it works with tokens, it is also meaningful to use the number of tokens as an indication of the difficulty of counting. Similarly, when the same letter appears on different tokens of a word, the LLM has not only to identify the presence of the letter in each token, but also to add them correctly, so it can also be an indication of the counting difficulty. The same happens when a letter appears more than once in a single token. To evaluate the difficulty of the counting we use the following features: 

\begin{enumerate}
    \item Number of letters in the word
    \item Number of token in the word
    \item Number of letters minus number of distinct letters in the word
    \item Number of letters  that appear in several tokens of the word minus number of distinct letters that appear in several tokens of the word
\end{enumerate}

For example, let us consider the word ``unbelievable'' that in GPT-4o is tokenized\footnote{see \url{https://platform.openai.com/tokenizer}} as: ``un'', ``bel'', ``ievable''. Then we will have 12 letters, 3 tokens, 8 distinct letters, 7 letters that appear in different tokens, and 3 distinct letters appearing in more than one token so having 12,3,4,4 for our four features.

\subsection{Words}

To keep the experiments manageable, we select 10,000 words randomly from the list of 1/3 million words in \cite{norvig2009natural}. Then for each of them we collect the frequency and difficulty features.

\subsection{LLMs}

In order to ensure that the results are representative of the current LLMs, we select both open and proprietary models from different companies and different sizes. In more detail, we evaluate: 

\begin{itemize}
    \item Three models from Meta: LLama3.1-8B, LLama3.1-70B and LLama3.2-11B \cite{llama3_1}, 
    \item Two from Mistral: Mistral-7B and Mixtral-8x7B \cite{Mistral}. 
    \item One model from Google: Gemma-2-9B \cite{team2024gemma2}.
    \item Two models from OpenAI: GPT-4o-mini and GPT-4o \cite{openai2023gpt4}.
\end{itemize}

\section{Evaluation Results}
\label{sec:Results}

\begin{figure}[H]
    \centering
    \includegraphics[width=0.9\linewidth]{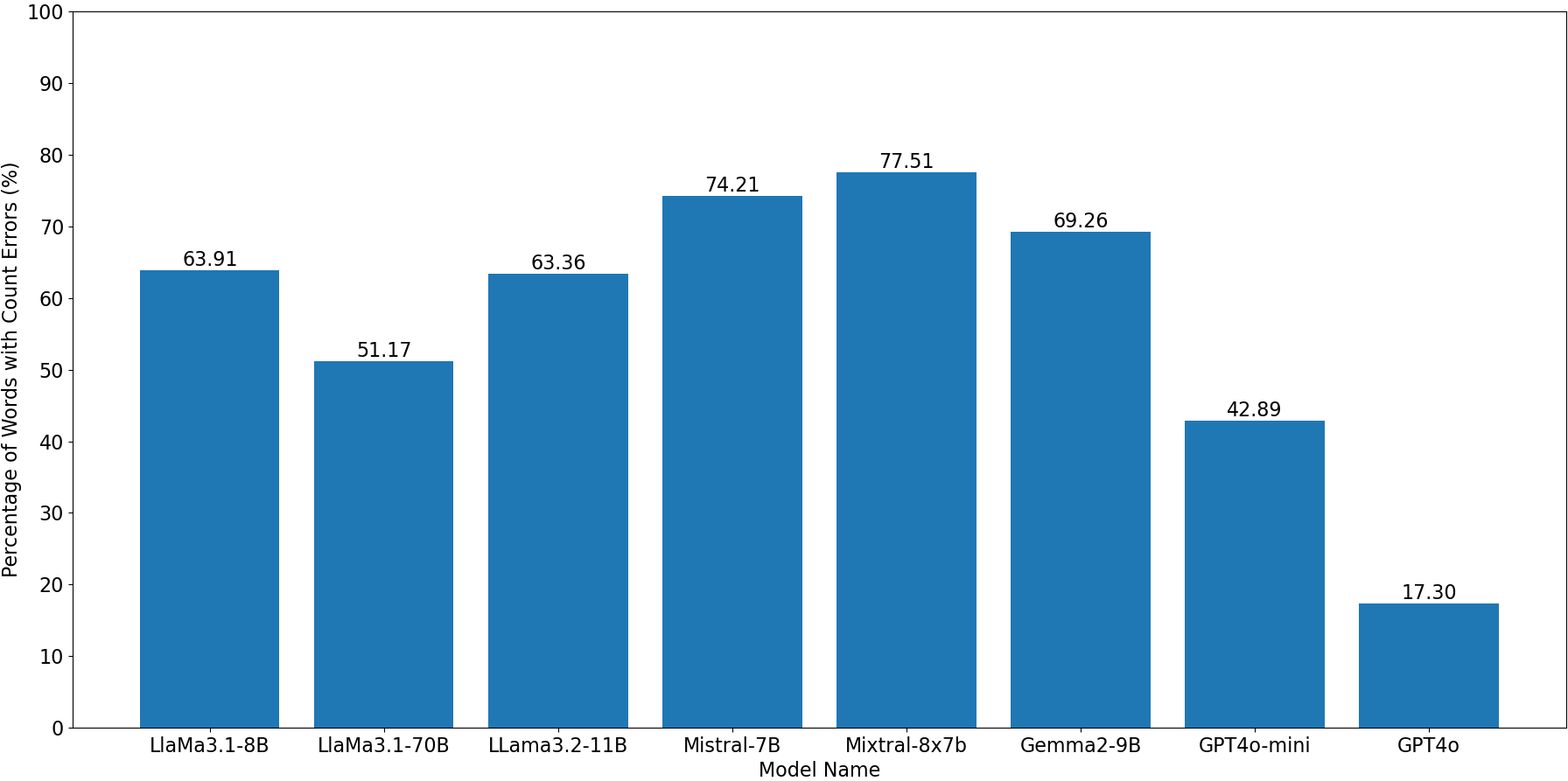}
    \caption{Percentage of words with errors on when counting letters for the different models}
    \label{fig:1}
\end{figure}

We first look at the overall results to understand if the latest versions of the LLMs are still unable to count the letters in a word\footnote{All the results and code are available at  \url{https://github.com/aMa2210/LLM_CounterLettersWithoutFT}}. Figure \ref{fig:1} shows the percentage of words with count errors per model. It can be observed that most models fail in more than half of the words and even the best model fails to correctly count the letters in 17\% of the words. Therefore, LLMs still struggle to count letters. 

Comparing the models, the largest and most powerful models evaluated, GPT-4o and GPT-4o-mini, show the best results with GPT-4o outperforming GPT-4o-mini. The best results for the open weights models evaluated are for LLama3.1-70B which is the largest of the open weights models evaluated. This suggests that larger models are better at counting letters. However, that is not the case for Mixtral-8x7B that has the worst results despite being the second largest open weights model evaluated. Instead, for the models in the 7B to 11B parameters range, performance is similar with errors in the range 63\% to 74\% for the four models evaluated: LLama3.1-8B, LLama3.2-11B, Mistral-7B and Gemma-2-9B.  
  
To try to understand whether frequency and difficulty features have an impact on errors, we analyze each group of features in the following subsections. 

\subsection{Frequency features}

\begin{figure}[H]
    \centering
    \includegraphics[width=0.9\linewidth]{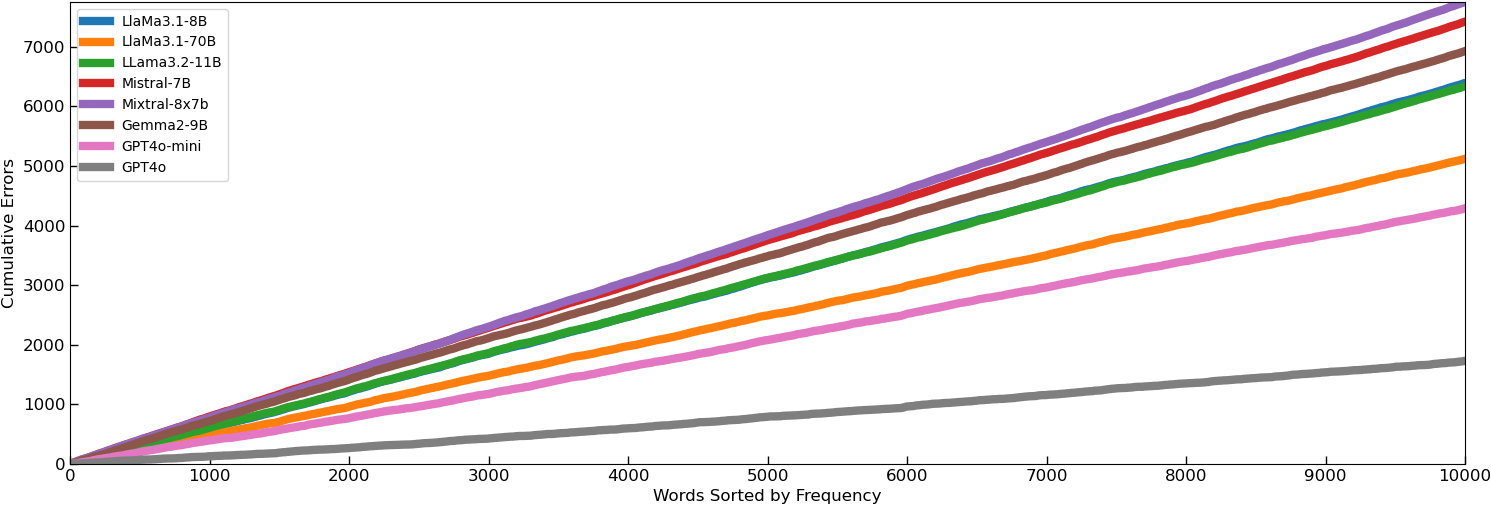}
    \caption{Distribution of cumulative error counts for words sorted by frequency}
    \label{fig:2}
\end{figure}

To analyze the impact of word frequency, we order the words by their frequency and plot the number of cumulative errors versus the number of words. Figure \ref{fig:2} shows the results. It can be observed that for all models, the errors increase linearly, which means that there is no dependence of errors with frequency. This is further confirmed when computing the Spearman correlation coefficient \cite{PearsonSpearman} between frequency and errors, which has values close to zero for all models. The values are provided in Table \ref{tab:Spearman}.

\begin{table}[!h]
    \centering
    \caption{Spearman Correlation Coefficients between Errors and Multiple Metrics}
    \resizebox{\textwidth}{!}{
        \begin{tabular}{c|c c c c c c c c}
            \hline
             & LlaMa3.1-8B & LlaMa3.1-70B & LLama3.2-11B & Mistral-7B & Mixtral-8x7B & Gemma2-9B & GPT4o-mini & GPT4o\\ \hline 
            Word Frequency & -0.04 & -0.04 & -0.03 & 0.02 & -0.02 & 0.01 & -0.04 & -0.05\\
            Average Frequency of Tokens & -0.12 & -0.09 & -0.12 & -0.15 & -0.11 & -0.09 & -0.08 & -0.09  \\ 
            Number of Letters & 0.55& 0.59& 0.56& 0.42& 0.40& 0.42& 0.58& 0.44\\
            Difference between Number of Letters and Distinct Letters & 0.72& 0.69& 0.61& 0.45& 0.48& 0.47& 0.66& 0.50\\
            Number of Tokens & 0.27 & 0.30 & 0.28 & 0.14 & 0.16 & 0.16 & 0.29 & 0.19\\
            Difference between Letters and Distinct Letters that appear on several tokens & 0.61 & 0.62 & 0.53 & 0.37 & 0.41 & 0.38 & 0.58 & 0.44\\
            \hline 
        \end{tabular}
        }
    \label{tab:Spearman}    
\end{table}

\begin{figure}[H]
    \centering
    \includegraphics[width=0.9\linewidth]{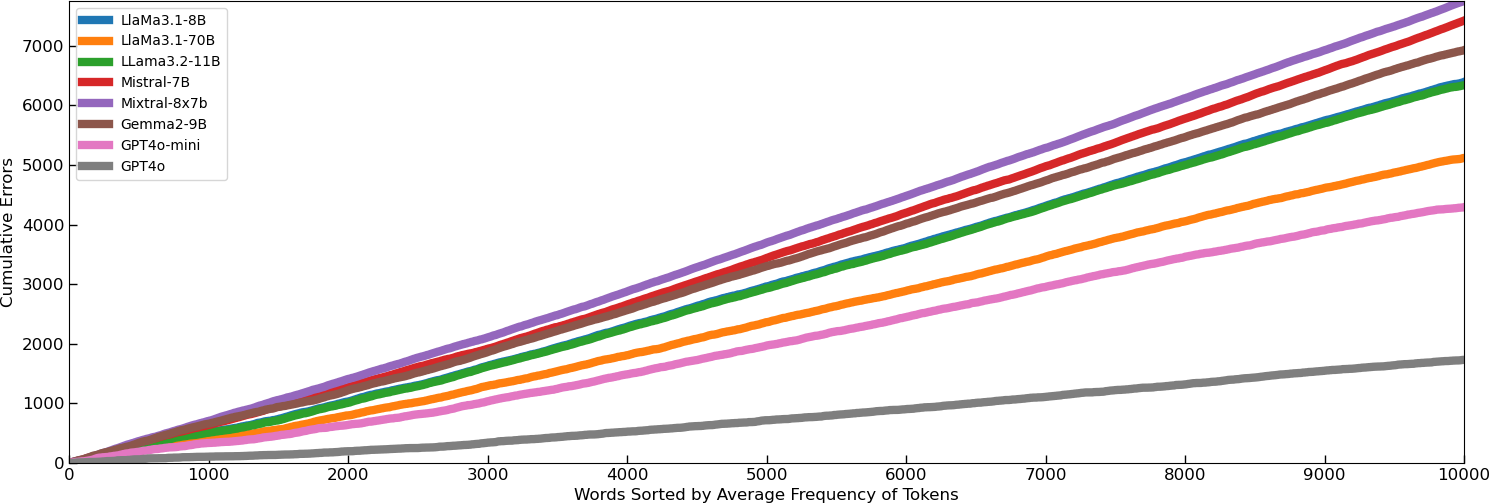}
    \caption{Distribution of cumulative error counts for words sorted by average frequency of tokens}
    \label{fig:3}
\end{figure}

Similar results are obtained when using the token frequency as seen in Figure \ref{fig:3}. The correlation coefficients are also close to zero as reported in Table \ref{tab:Spearman}. These results suggest that the frequency of the word or its tokens does not have an impact on the counting errors of LLMs. This is surprising and implies that the failures in letter counting are not related to the times that the word is expected to appear in the training set of the LLM.

\begin{figure}[H]
    \centering
    \includegraphics[width=0.9\linewidth]{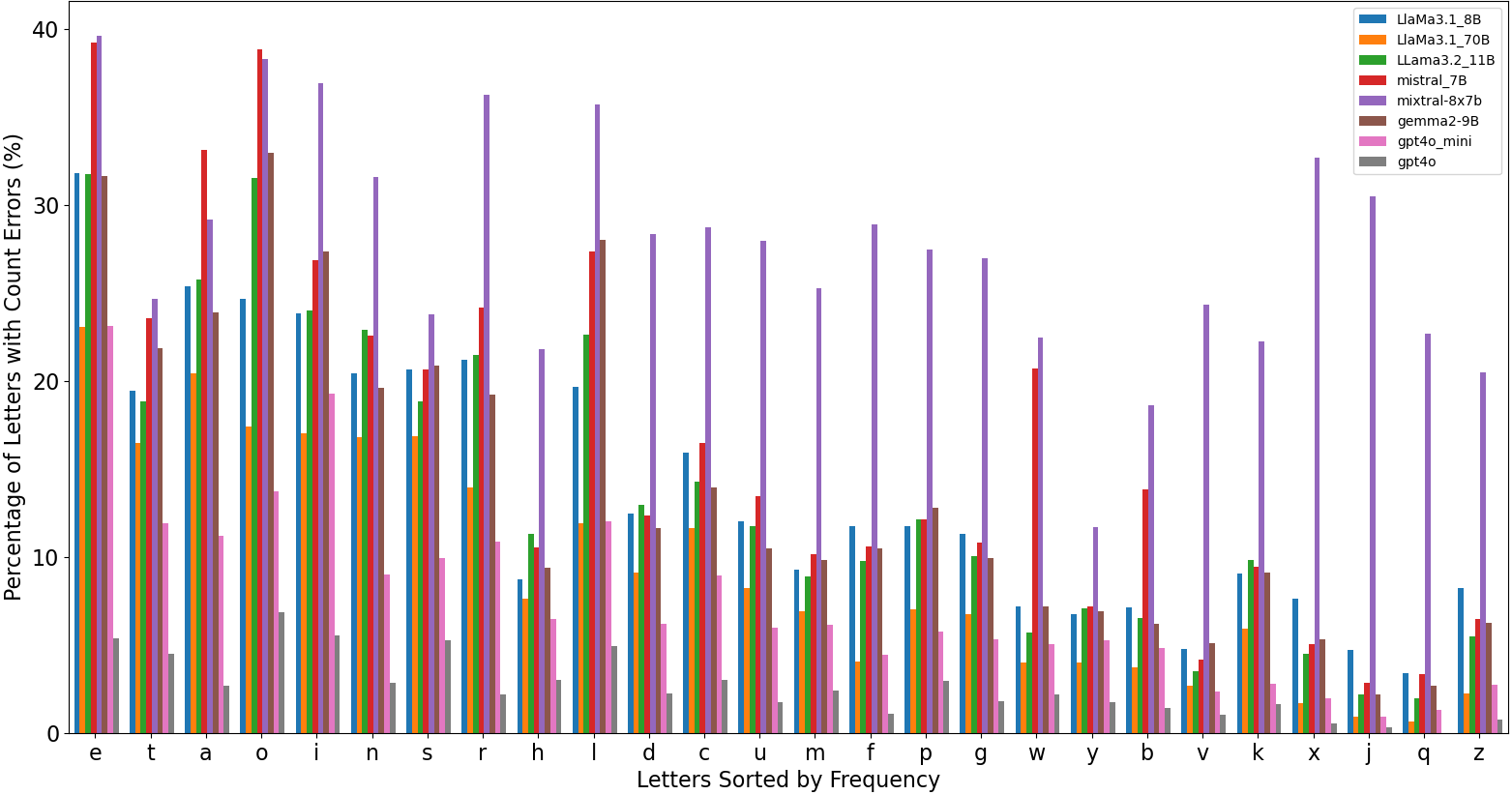}
    \caption{Relationship between letter counting errors and letter frequency}
    \label{fig:4}
\end{figure}

Finally, we consider the errors at the letter level and their dependency with letter frequency. To do so, we order the letters by frequency\footnote{We use the ordering in \url{https://norvig.com/mayzner.html}} and plot their counting error rates in Figure \ref{fig:4}. A strong correlation between frequency and counting errors can be observed for all models except Mixtral-8x7b which does not show a clear trend. The correlation of counting errors with letter frequency is not intuitive, as it seems that if anything, frequency should help. As we will see in the following subsection there is an explanation for this behaviour.  

%To understand the correlation of errors with letter frequency we compute for each letter the frequency with which it appears more than once in a word, so if it appears on 100 words, in how many of those it does so more than once. The results versus this multiple occurrence frequency are shown in Figure \ref{fig:5}.  It can be observed that in this case the correlation is even better. This suggests, that it is the appearance of the letter multiple time what is related to the errors. 

%\begin{figure}[H]
%    \centering
%    \includegraphics[width=0.9\linewidth]{Figures/Letter_Mul_Occur_Frequency.png}
%    \caption{Relationship between letter counting errors and frequency of multiple occurrence}
%    \label{fig:5}
%\end{figure}

\subsection{Count difficulty features}

\begin{figure}[H]
    \centering
    \includegraphics[width=0.9\linewidth]{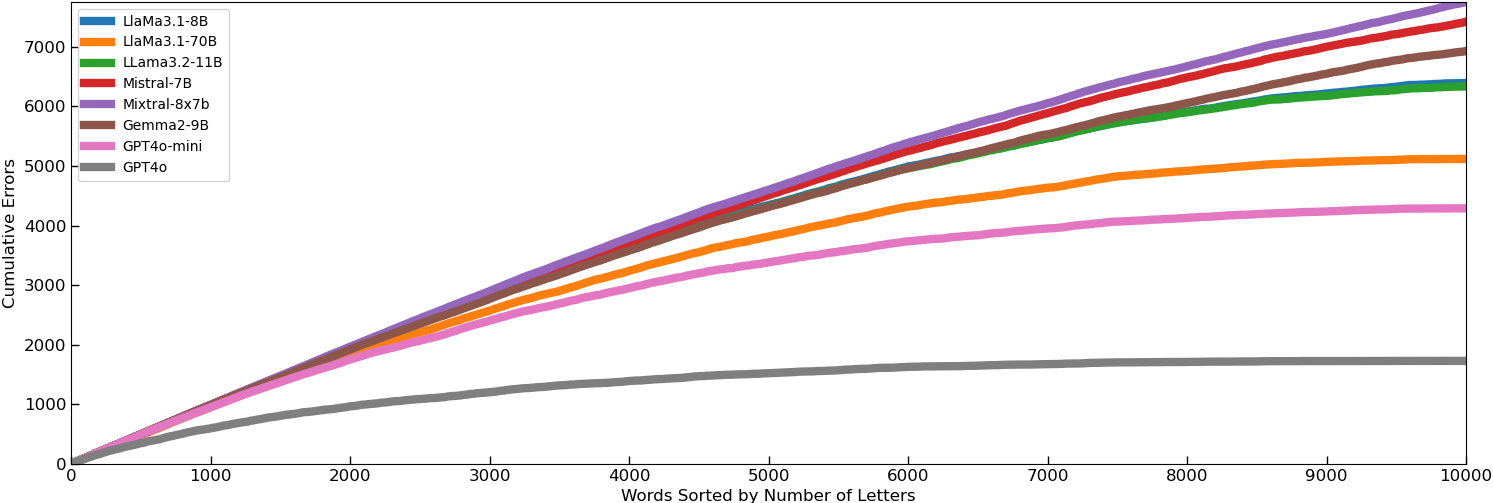}
    \caption{Distribution of cumulative error counts for words sorted by letter numbers in descending order}
    \label{fig:6}
\end{figure}

\begin{figure}[H]
    \centering
    \includegraphics[width=0.9\linewidth]{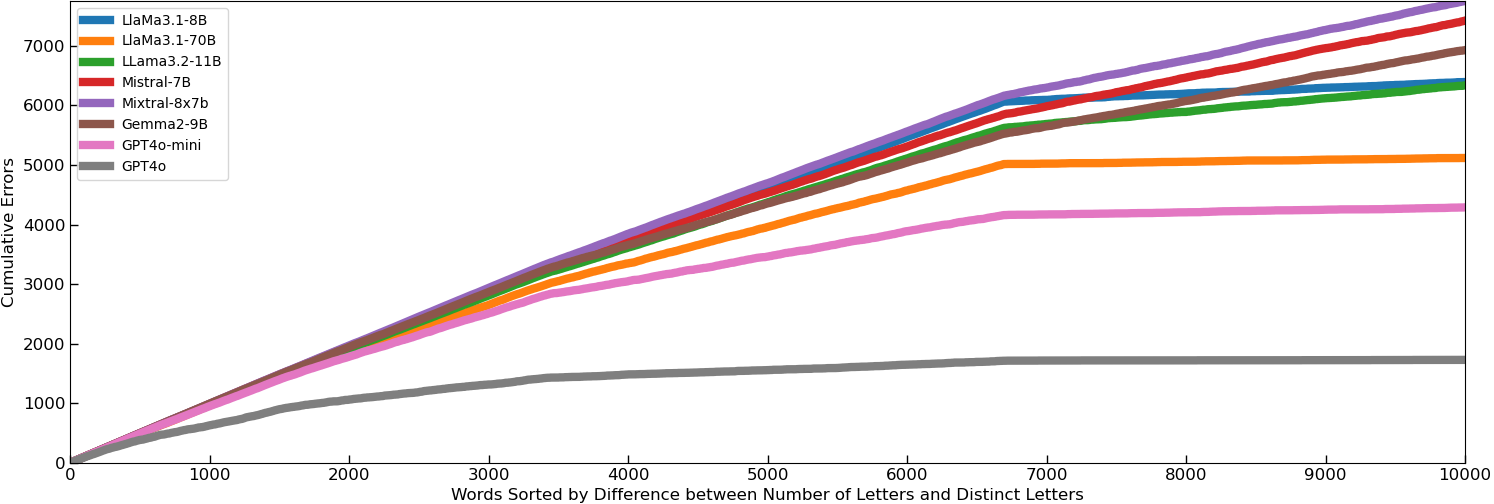}
    \caption{Distribution of cumulative error counts for words sorted by difference of number of letters and distinct letters in descending order}
    \label{fig:7}
\end{figure}

Figures \ref{fig:6} and \ref{fig:7} show the cumulative word errors as a function of the number of letters, and the difference between the number of letters and distinct letters when words are ordered by these parameters in descending order, respectively. It can be seen that most errors occur with the first words so errors increase with the two features. This is expected as the more letters to count, the harder the problem. Looking closer at the results, the increase in the errors is greater for the difference in counts. This effect is more significant for models that have fewer errors for which errors stabilize quickly. This is consistent with Table \ref{tab:Spearman} on which the correlation coefficients are also larger for the difference of letters and distinct letters than for the number of letters. 

\begin{figure}[H]
    \centering
    \includegraphics[width=0.9\linewidth]{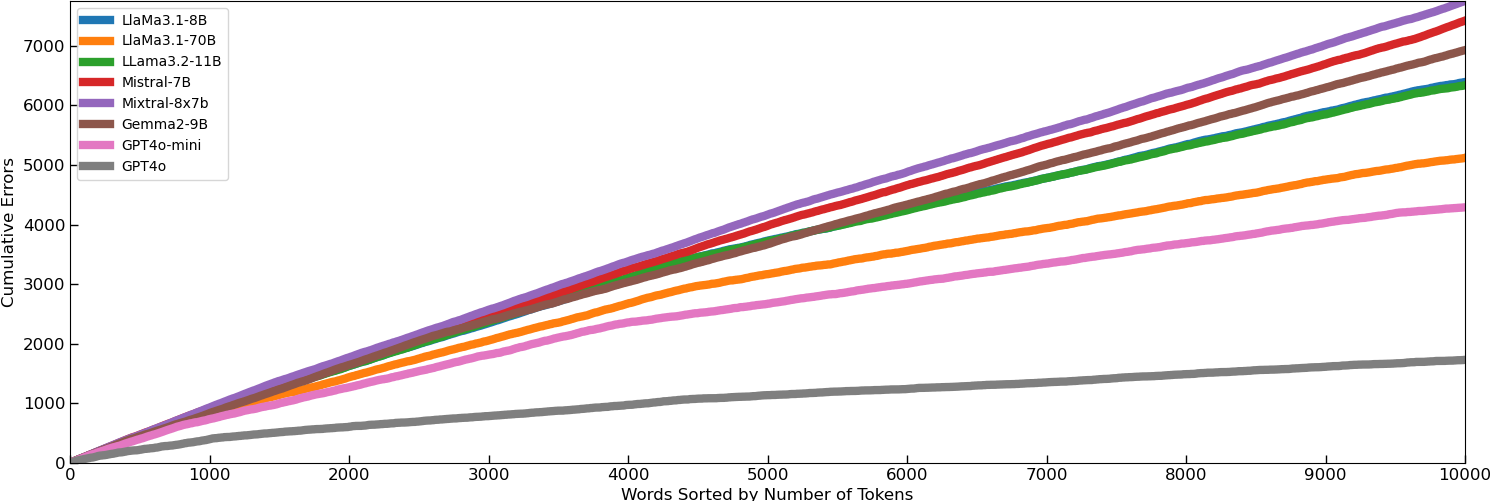}
    \caption{Distribution of cumulative error counts for words sorted by number of tokens in descending order}
    \label{fig:8}
\end{figure}

\begin{figure}[H]
    \centering
    \includegraphics[width=0.9\linewidth]{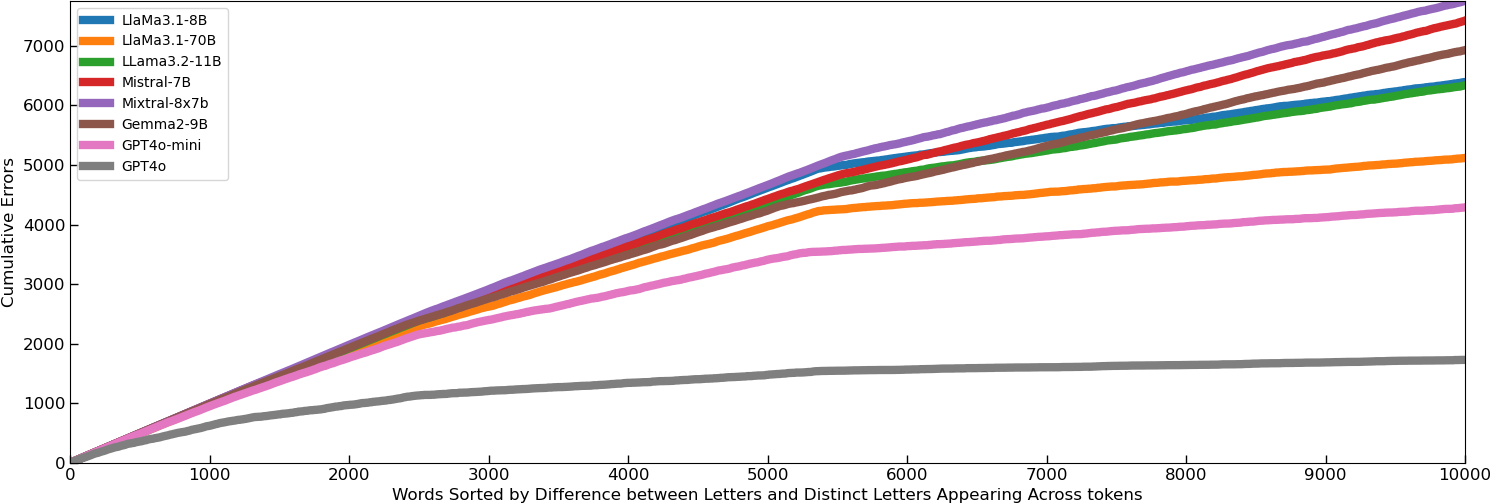}
    \caption{Distribution of cumulative error counts for words sorted by difference of number of letters and distinct letters appearing across tokens  in descending order}
    \label{fig:9}
\end{figure}

Now we look at the token-related features in figures \ref{fig:8} and \ref{fig:9}. It can be seen that again there is a correlation with errors that is stronger for the difference of letters and distinct letters appearing on several tokens as with letters. Compared with the same metrics for letters, it can be seen in table \ref{tab:Spearman} that the correlation with errors is lower for tokens. This suggests that tokenization is not the fundamental issue that causes LLMs to struggle when counting letters.

\begin{figure}[H]
    \centering
    \includegraphics[width=0.9\linewidth]{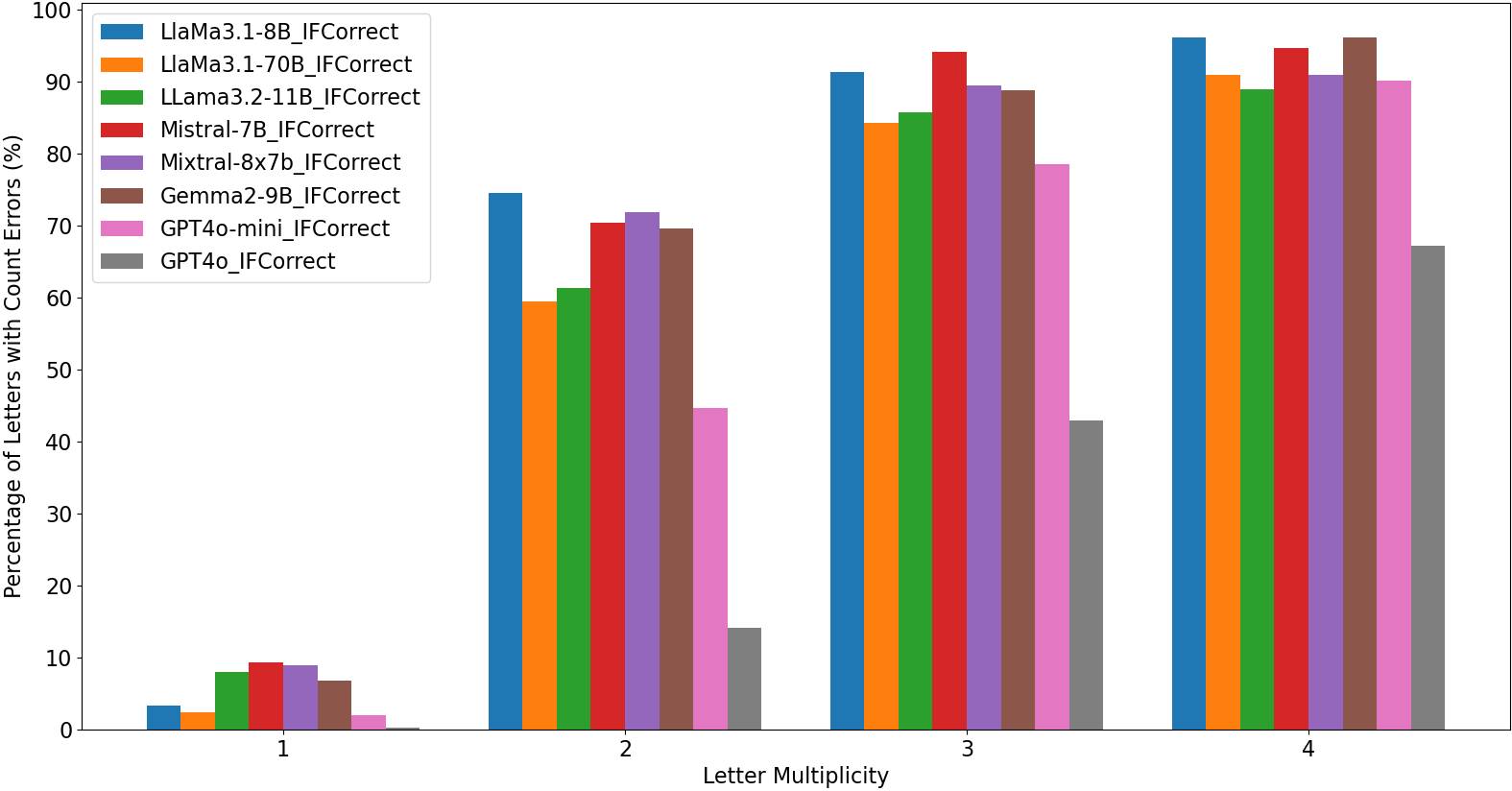}
    \caption{Percentage of errors when counting a letter versus the multiplicity of the letter in the word}
    \label{fig:10}
\end{figure}

These results suggest that errors correlate strongly with words that have letters with a multiplicity greater than one. Therefore, it is interesting to check how the letter multiplicity affects the LLM performance. To do so, in figure \ref{fig:10} we plot the percentage of errors when counting a letter as a function of the letter multiplicity. So one means the letter appears only once in the word, two that it appears twice, and so on. It can be observed that the percentage of errors increases very significantly when the letter count is larger than one for all the models. This is consistent with the results presented in the previous figures that show the strongest dependency with the number of letters minus the number of distinct letters, which increases when letters appear multiple times on a word. The correlation of errors with letter multiplicity also explains the relation between the frequency of letters and the errors observed in figure \ref{fig:4} as, the larger the frequency, the higher the probability of having words on which the letter appears several times. 

 %\textcolor{red}{I have only included the results shown in the figure below. The complete results can be found in Figures/CountErrors\_MostLetterNum\_all.png. When the multiplicity is greater than 4, the sample sizes are all less than 10 (5: 9; 6: 2; 8: 1).}

%Finally, to check the impact of tokenization, we compute the same metric but only for letters that appear only in a single token in a word. In this case, the counting for the letter involves only that token, so if tokenization was the fundamental issue we would expect similar results to those of figure \ref{fig:10}. The results are presented in figure \ref{fig:11}. It can be observed that the errors also increase with letter multiplicity but differently from figure \ref{fig:10} which suggests that counting the letters in a single token is only part of the problem.

\begin{figure}[H]
    \centering
    \includegraphics[width=0.9\linewidth]{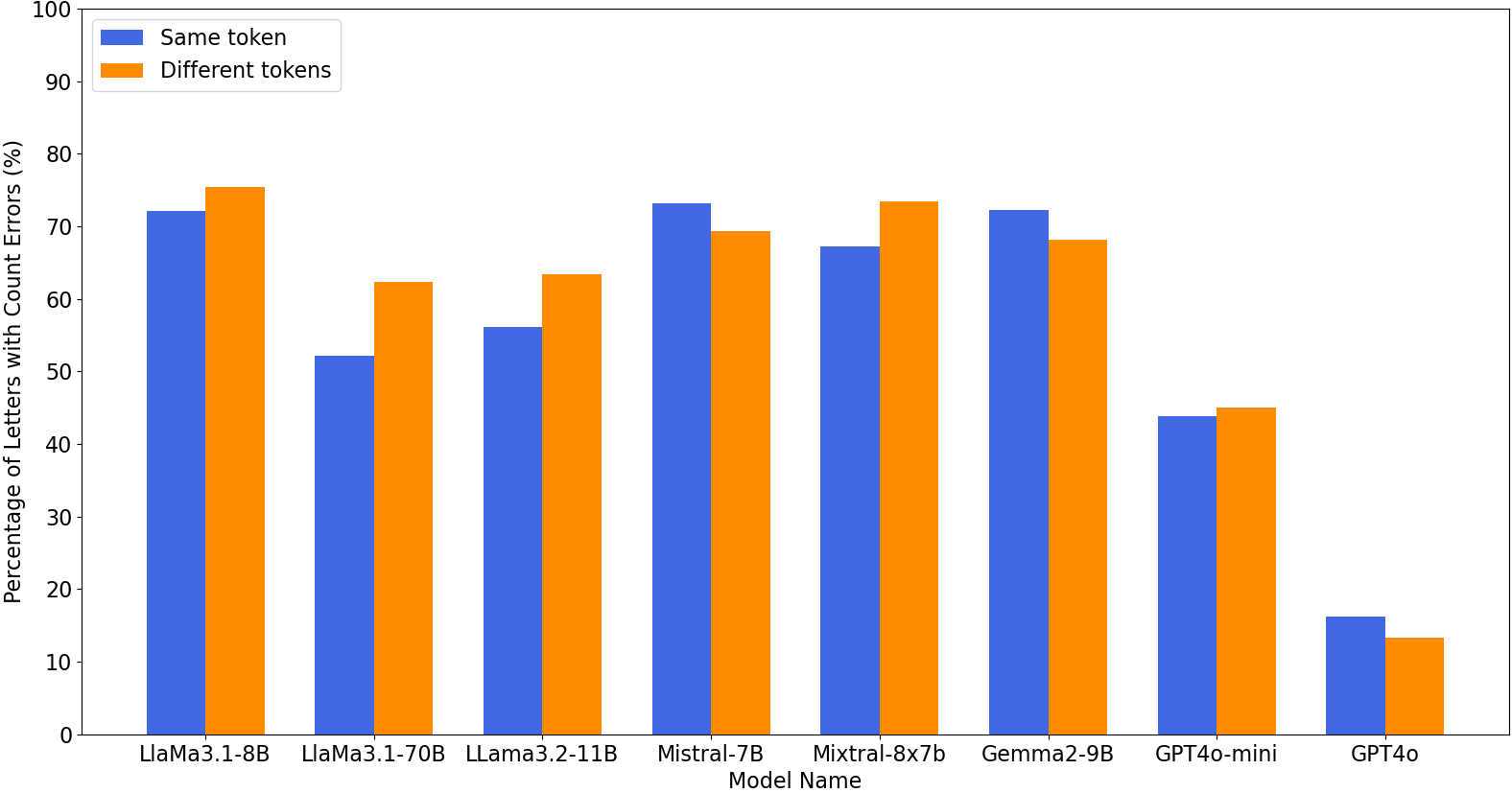}
    \caption{Percentage of errors when counting a letter that appears twice on a word when a) the two occurrences are in the same token and b) in different tokens}
    \label{fig:11}
\end{figure}

To check the impact of tokenization, we compute the same metric but only for letters that appear exactly twice in a word in a) the same token and b) two different tokens. The results for each model are presented in figure \ref{fig:11}. It can be seen that the percentage of errors is similar in both cases. From figure \ref{fig:10} we see that LLMs correctly identify letters that appear only once in a token, therefore the failures when the letter appears in two tokens seem to be related to the counting of the letters and not to a limitation in identifying the letters in the tokens. This suggests that tokenization is not the main problem when counting letters. 

\begin{figure}[H]
    \centering
    \includegraphics[width=0.9\linewidth]{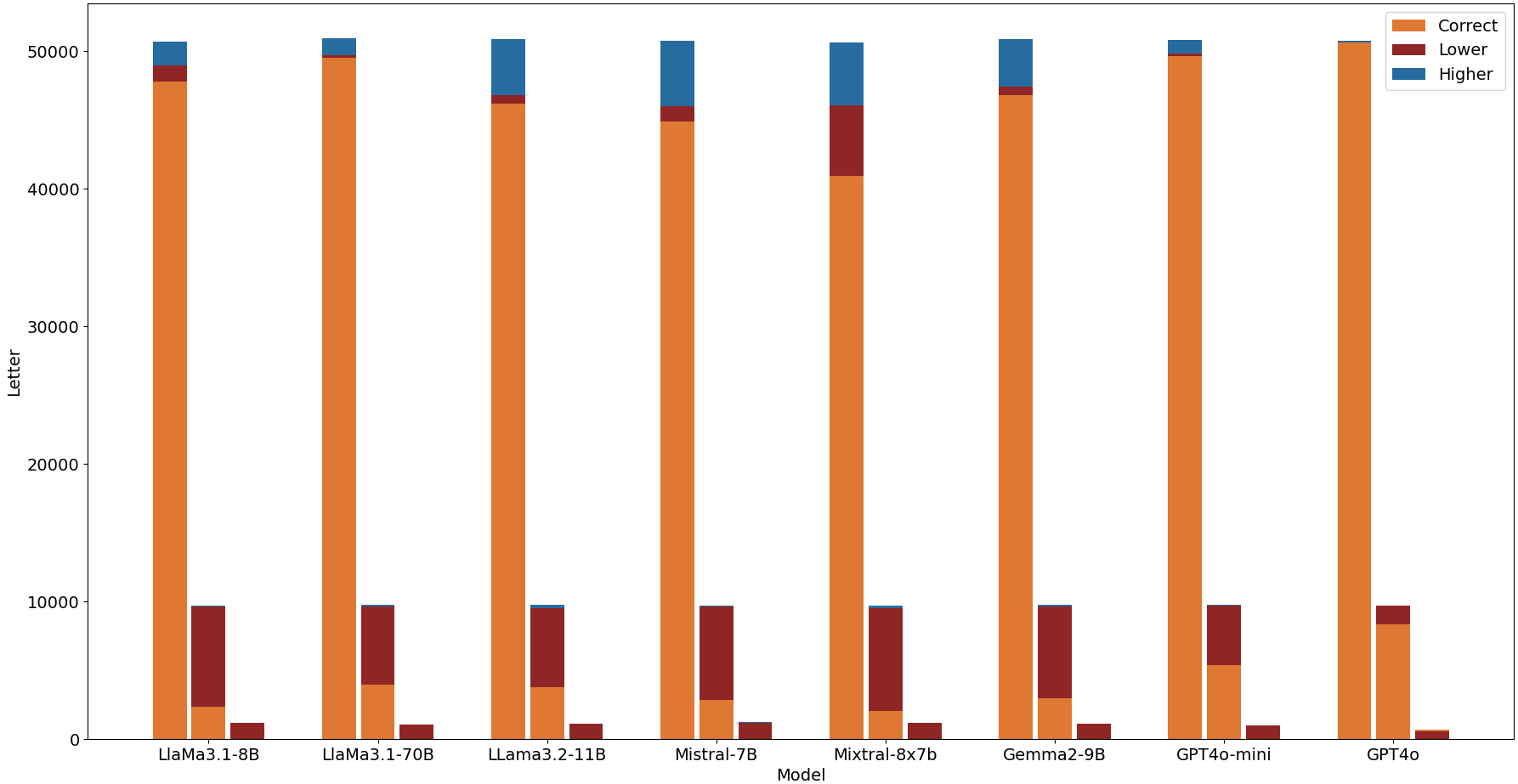}
    \caption{Results per model and letter multiplicity (being each column in each model 1,2, and 3 respectively) with a breakdown of errors into lower and higher than the real value.}
    \label{fig:12}
\end{figure}

Finally, we take a closer look at the errors to see whether the models tend to produce lower or higher counts than the real value. To do so for each letter multiplicity, we breakdown the errors in correct, lower and higher and show the results per model in figure \ref{fig:12}. It can be seen that when the letter multiplicity is one, most errors are due to the models responding with values larger than one. Instead, for multiplicities of two and three most errors are caused by responses with lower values. The exception is Mixtral-8x7B that for multiplicity of one has more answers with zero than with values higher than one. A detailed analysis of the results for each letter is provided in the Appendix.

\begin{figure}[H]
    \centering
    \includegraphics[width=0.9\linewidth]{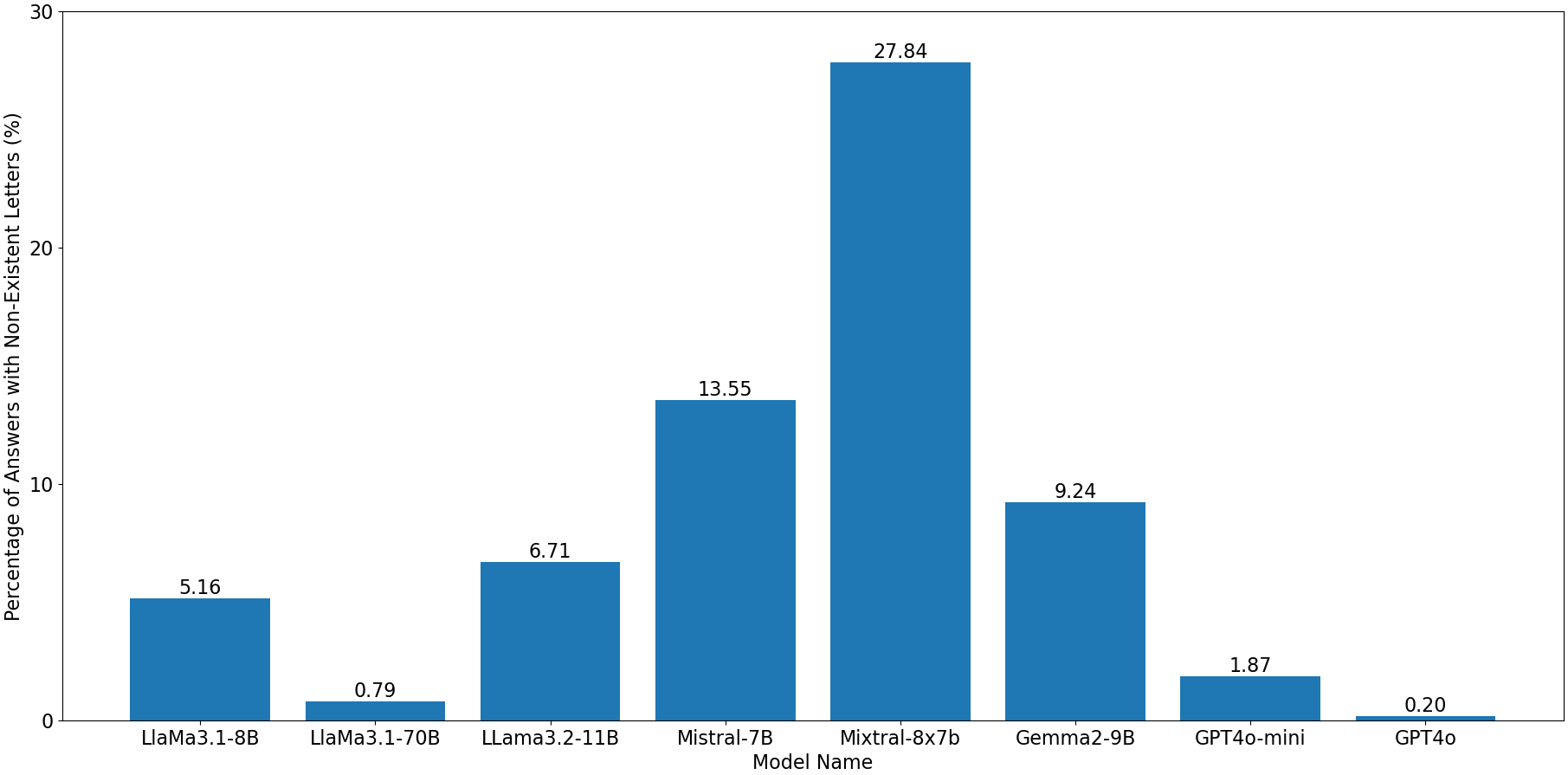}
    \caption{Percentage of words for which the models respond with count values larger than zero for letters that do not appear in the word}
    \label{fig:13}
\end{figure}

To complete the analysis, we look at the percentage of words for which the models report counts larger than zero for letters that are not in the word. The results are shown in figure \ref{fig:13}. The models do not tend to include letters that do not appear in the word, which proves that the models are capable of detecting the letters, and their limitation is counting them. It can be observed that again Mixtral-8x7B has the worse results and that larger models for Llama and for GPT4o reduce the number of errors significantly. 

\subsection{Discussion}

The analysis of the results shows that models concentrate most of their errors when counting letters that have a multiplicity greater than one. This can be due to a number of reasons. The first one is that models may be unable to identify letters with multiplicity larger than one in a token. For example, in the tokenizer used by GPT-4o, ``berry'' is a token, if the model is unable to identify that it has two ``r'', then any word that contains ``berry'' is likely to have counting errors. The second one could be that the model can identify the letters in each token of a word but fails to add them to compute the counts for the word. Finally, it could be that since most letters have counts of either zero or one in a word, the models tend to answer with those values, as it is what they have seen mostly during training.  The results suggest that the main problem to correctly count letters in a word lies in the counting itself and that tokenization does not play a fundamental role and word or token frequency have no impact on the result.

%\textcolor{red}{Finally, it is of interest to understand if models that have been optimized for reasoning like o1 from OpenAI or QwQ from Alibaba can count letters. To do this, we repeated the test with QwQ-32B\footnote{\url{https://huggingface.co/spaces/Qwen/QwQ-32B-preview}} and the results show TBD.}

%To check if the first reason could be causing the errors, we have evaluated the errors on words that are mapped to a single token for two of the models, GPT-4o and LLama3.1-8B. The results show that for errors with letter count of two, the percentage on single token words is TBD while to words that are made of several tokens is TBD. 

\section{Conclusion}
\label{sec:Conclusion}

This work has explored the factors that influence the errors made by LLMs when counting letters in a word by evaluating eight different LLMs over 10,000 words. The results show that there is no dependency of the errors with the word or token frequency and that errors are mostly related to the difficulty of the counting in terms of the number of letters and especially in terms of the number of letters with multiplicity larger than one. This behavior is consistent across all the models evaluated and suggests that there is a general limitation of LLMs to count letters that appear multiple times in a word.

%%
%% The acknowledgments section is defined using the "acks" environment
%% (and NOT an unnumbered section). This ensures the proper
%% identification of the section in the article metadata, and the
%% consistent spelling of the heading.
\begin{acks}

This work was supported by the FUN4DATE (PID2022-136684OB-C22) project funded by the Spanish Agencia Estatal de Investigacion (AEI) 10.13039/501100011033, by the Chips Act Joint Undertaking project SMARTY (Grant no. 101140087) and by the OpenAI API research access program.

\end{acks}

%%
%% The next two lines define the bibliography style to be used, and
%% the bibliography file.

\bibliographystyle{ACM-Reference-Format}

\bibliography{LetterCount}

%%% -*-BibTeX-*-
%%% Do NOT edit. File created by BibTeX with style
%%% ACM-Reference-Format-Journals [18-Jan-2012].

\begin{thebibliography}{25}

%%% ====================================================================
%%% NOTE TO THE USER: you can override these defaults by providing
%%% customized versions of any of these macros before the \bibliography
%%% command.  Each of them MUST provide its own final punctuation,
%%% except for \shownote{}, \showDOI{}, and \showURL{}.  The latter two
%%% do not use final punctuation, in order to avoid confusing it with
%%% the Web address.
%%%
%%% To suppress output of a particular field, define its macro to expand
%%% to an empty string, or better, \unskip, like this:
%%%
%%% \newcommand{\showDOI}[1]{\unskip}   % LaTeX syntax
%%%
%%% \def \showDOI #1{\unskip}           % plain TeX syntax
%%%
%%% ====================================================================

\ifx \showCODEN    \undefined \def \showCODEN     #1{\unskip}     \fi
\ifx \showDOI      \undefined \def \showDOI       #1{#1}\fi
\ifx \showISBNx    \undefined \def \showISBNx     #1{\unskip}     \fi
\ifx \showISBNxiii \undefined \def \showISBNxiii  #1{\unskip}     \fi
\ifx \showISSN     \undefined \def \showISSN      #1{\unskip}     \fi
\ifx \showLCCN     \undefined \def \showLCCN      #1{\unskip}     \fi
\ifx \shownote     \undefined \def \shownote      #1{#1}          \fi
\ifx \showarticletitle \undefined \def \showarticletitle #1{#1}   \fi
\ifx \showURL      \undefined \def \showURL       {\relax}        \fi
% The following commands are used for tagged output and should be
% invisible to TeX
\providecommand\bibfield[2]{#2}
\providecommand\bibinfo[2]{#2}
\providecommand\natexlab[1]{#1}
\providecommand\showeprint[2][]{arXiv:#2}

\bibitem[Baroni et~al\mbox{.}(2011)]%
        {baroni2011alphabetic}
\bibfield{author}{\bibinfo{person}{Antonio Baroni} {et~al\mbox{.}}} \bibinfo{year}{2011}\natexlab{}.
\newblock \showarticletitle{Alphabetic vs. non-alphabetic writing: Linguistic fit and natural tendencies}.
\newblock \bibinfo{journal}{\emph{Rivista di Linguistica}} \bibinfo{volume}{23}, \bibinfo{number}{2} (\bibinfo{year}{2011}), \bibinfo{pages}{127--159}.
\newblock


\bibitem[Barratt-Pugh and Rohl(2020)]%
        {barratt2020literacy}
\bibfield{author}{\bibinfo{person}{Caroline Barratt-Pugh} {and} \bibinfo{person}{Mary Rohl}.} \bibinfo{year}{2020}\natexlab{}.
\newblock \bibinfo{booktitle}{\emph{Literacy learning in the early years}}.
\newblock \bibinfo{publisher}{Routledge}.
\newblock


\bibitem[Borges(1962)]%
        {borges1962funes}
\bibfield{author}{\bibinfo{person}{Jorge~Luis Borges}.} \bibinfo{year}{1962}\natexlab{}.
\newblock \bibinfo{booktitle}{\emph{Funes, the memorious}}.
\newblock \bibinfo{publisher}{na}.
\newblock


\bibitem[Dubey et~al\mbox{.}(2024)]%
        {llama3_1}
\bibfield{author}{\bibinfo{person}{Abhimanyu Dubey}, \bibinfo{person}{Abhinav Jauhri}, \bibinfo{person}{Abhinav Pandey}, \bibinfo{person}{Abhishek Kadian}, \bibinfo{person}{Ahmad Al-Dahle}, \bibinfo{person}{Aiesha Letman}, \bibinfo{person}{Akhil Mathur}, \bibinfo{person}{Alan Schelten}, \bibinfo{person}{Amy Yang}, \bibinfo{person}{Angela Fan}, {et~al\mbox{.}}} \bibinfo{year}{2024}\natexlab{}.
\newblock \showarticletitle{The llama 3 herd of models}.
\newblock \bibinfo{journal}{\emph{arXiv preprint arXiv:2407.21783}} (\bibinfo{year}{2024}).
\newblock


\bibitem[Edman et~al\mbox{.}(2024)]%
        {edman-etal-2024-cute}
\bibfield{author}{\bibinfo{person}{Lukas Edman}, \bibinfo{person}{Helmut Schmid}, {and} \bibinfo{person}{Alexander Fraser}.} \bibinfo{year}{2024}\natexlab{}.
\newblock \showarticletitle{{CUTE}: Measuring {LLM}s{'} Understanding of Their Tokens}. In \bibinfo{booktitle}{\emph{Proceedings of the 2024 Conference on Empirical Methods in Natural Language Processing}}, \bibfield{editor}{\bibinfo{person}{Yaser Al-Onaizan}, \bibinfo{person}{Mohit Bansal}, {and} \bibinfo{person}{Yun-Nung Chen}} (Eds.). \bibinfo{publisher}{Association for Computational Linguistics}, \bibinfo{address}{Miami, Florida, USA}, \bibinfo{pages}{3017--3026}.
\newblock
\urldef\tempurl%
\url{https://doi.org/10.18653/v1/2024.emnlp-main.177}
\showDOI{\tempurl}


\bibitem[et~al(2023)]%
        {Mistral}
\bibfield{author}{\bibinfo{person}{Albert Q.~Jiang et al}.} \bibinfo{year}{2023}\natexlab{}.
\newblock \bibinfo{title}{Mistral 7B}.
\newblock
\newblock
\showeprint[arxiv]{2310.06825}~[cs.CL]


\bibitem[Helford(1988)]%
        {funes_analysis}
\bibfield{author}{\bibinfo{person}{Elyce~Rae Helford}.} \bibinfo{year}{1988}\natexlab{}.
\newblock \showarticletitle{Language and Memory in Borges'" Funes, The Memorious"}.
\newblock \bibinfo{journal}{\emph{Iowa Journal of Literary Studies}} \bibinfo{volume}{9}, \bibinfo{number}{1} (\bibinfo{year}{1988}).
\newblock


\bibitem[Hendrycks et~al\mbox{.}(2021)]%
        {Mathmeasuring}
\bibfield{author}{\bibinfo{person}{Dan Hendrycks}, \bibinfo{person}{Collin Burns}, \bibinfo{person}{Saurav Kadavath}, \bibinfo{person}{Akul Arora}, \bibinfo{person}{Steven Basart}, \bibinfo{person}{Eric Tang}, \bibinfo{person}{Dawn Song}, {and} \bibinfo{person}{Jacob Steinhardt}.} \bibinfo{year}{2021}\natexlab{}.
\newblock \bibinfo{title}{Measuring Mathematical Problem Solving With the MATH Dataset}.
\newblock
\newblock
\showeprint[arxiv]{2103.03874}~[cs.LG]


\bibitem[Minaee et~al\mbox{.}(2024)]%
        {llm_general_survey}
\bibfield{author}{\bibinfo{person}{Shervin Minaee}, \bibinfo{person}{Tomas Mikolov}, \bibinfo{person}{Narjes Nikzad}, \bibinfo{person}{Meysam Chenaghlu}, \bibinfo{person}{Richard Socher}, \bibinfo{person}{Xavier Amatriain}, {and} \bibinfo{person}{Jianfeng Gao}.} \bibinfo{year}{2024}\natexlab{}.
\newblock \showarticletitle{Large language models: A survey}.
\newblock \bibinfo{journal}{\emph{arXiv preprint arXiv:2402.06196}} (\bibinfo{year}{2024}).
\newblock


\bibitem[Myers et~al\mbox{.}(2013)]%
        {PearsonSpearman}
\bibfield{author}{\bibinfo{person}{Jerome~L Myers}, \bibinfo{person}{Arnold~D Well}, {and} \bibinfo{person}{Robert~F Lorch~Jr}.} \bibinfo{year}{2013}\natexlab{}.
\newblock \bibinfo{booktitle}{\emph{Research design and statistical analysis}}.
\newblock \bibinfo{publisher}{Routledge}.
\newblock


\bibitem[Norvig(2009)]%
        {norvig2009natural}
\bibfield{author}{\bibinfo{person}{Peter Norvig}.} \bibinfo{year}{2009}\natexlab{}.
\newblock \showarticletitle{Natural language corpus data}.
\newblock \bibinfo{journal}{\emph{Beautiful data}} (\bibinfo{year}{2009}), \bibinfo{pages}{219--242}.
\newblock


\bibitem[OpenAI(2023)]%
        {openai2023gpt4}
\bibfield{author}{\bibinfo{person}{OpenAI}.} \bibinfo{year}{2023}\natexlab{}.
\newblock \bibinfo{title}{GPT-4 Technical Report}.
\newblock
\newblock
\showeprint[arxiv]{2303.08774}~[cs.CL]


\bibitem[Petrov et~al\mbox{.}(2023)]%
        {NEURIPS2023_74bb24dc}
\bibfield{author}{\bibinfo{person}{Aleksandar Petrov}, \bibinfo{person}{Emanuele La~Malfa}, \bibinfo{person}{Philip Torr}, {and} \bibinfo{person}{Adel Bibi}.} \bibinfo{year}{2023}\natexlab{}.
\newblock \showarticletitle{Language Model Tokenizers Introduce Unfairness Between Languages}. In \bibinfo{booktitle}{\emph{Advances in Neural Information Processing Systems}}, \bibfield{editor}{\bibinfo{person}{A.~Oh}, \bibinfo{person}{T.~Naumann}, \bibinfo{person}{A.~Globerson}, \bibinfo{person}{K.~Saenko}, \bibinfo{person}{M.~Hardt}, {and} \bibinfo{person}{S.~Levine}} (Eds.), Vol.~\bibinfo{volume}{36}. \bibinfo{publisher}{Curran Associates, Inc.}, \bibinfo{pages}{36963--36990}.
\newblock
\urldef\tempurl%
\url{https://proceedings.neurips.cc/paper_files/paper/2023/file/74bb24dca8334adce292883b4b651eda-Paper-Conference.pdf}
\showURL{%
\tempurl}


\bibitem[Rahman et~al\mbox{.}(2024)]%
        {TokenizerLanguages}
\bibfield{author}{\bibinfo{person}{Abrar Rahman}, \bibinfo{person}{Garry Bowlin}, \bibinfo{person}{Binit Mohanty}, {and} \bibinfo{person}{Sean McGunigal}.} \bibinfo{year}{2024}\natexlab{}.
\newblock \showarticletitle{Towards Linguistically-Aware and Language-Independent Tokenization for Large Language Models (LLMs)}.
\newblock \bibinfo{journal}{\emph{arXiv preprint arXiv:2410.03568}} (\bibinfo{year}{2024}).
\newblock


\bibitem[Rajaraman et~al\mbox{.}(2024)]%
        {TokenizerTheo}
\bibfield{author}{\bibinfo{person}{Nived Rajaraman}, \bibinfo{person}{Jiantao Jiao}, {and} \bibinfo{person}{Kannan Ramchandran}.} \bibinfo{year}{2024}\natexlab{}.
\newblock \showarticletitle{Toward a Theory of Tokenization in LLMs}.
\newblock \bibinfo{journal}{\emph{arXiv preprint arXiv:2404.08335}} (\bibinfo{year}{2024}).
\newblock


\bibitem[Sennrich et~al\mbox{.}(2016)]%
        {Tokenizer1}
\bibfield{author}{\bibinfo{person}{Rico Sennrich}, \bibinfo{person}{Barry Haddow}, {and} \bibinfo{person}{Alexandra Birch}.} \bibinfo{year}{2016}\natexlab{}.
\newblock \showarticletitle{Neural Machine Translation of Rare Words with Subword Units}. In \bibinfo{booktitle}{\emph{Proceedings of the 54th Annual Meeting of the Association for Computational Linguistics (Volume 1: Long Papers)}}. \bibinfo{pages}{1715--1725}.
\newblock


\bibitem[Shin and Kaneko(2024)]%
        {CountLetters3}
\bibfield{author}{\bibinfo{person}{Andrew Shin} {and} \bibinfo{person}{Kunitake Kaneko}.} \bibinfo{year}{2024}\natexlab{}.
\newblock \showarticletitle{Large language models lack understanding of character composition of words}.
\newblock \bibinfo{journal}{\emph{arXiv preprint arXiv:2405.11357}} (\bibinfo{year}{2024}).
\newblock


\bibitem[Team et~al\mbox{.}(2024)]%
        {team2024gemma2}
\bibfield{author}{\bibinfo{person}{Gemma Team}, \bibinfo{person}{Morgane Riviere}, \bibinfo{person}{Shreya Pathak}, \bibinfo{person}{Pier~Giuseppe Sessa}, \bibinfo{person}{Cassidy Hardin}, \bibinfo{person}{Surya Bhupatiraju}, \bibinfo{person}{L{\'e}onard Hussenot}, \bibinfo{person}{Thomas Mesnard}, \bibinfo{person}{Bobak Shahriari}, \bibinfo{person}{Alexandre Ram{\'e}}, {et~al\mbox{.}}} \bibinfo{year}{2024}\natexlab{}.
\newblock \showarticletitle{Gemma 2: Improving open language models at a practical size}.
\newblock \bibinfo{journal}{\emph{arXiv e-prints}} (\bibinfo{year}{2024}), \bibinfo{pages}{arXiv--2408}.
\newblock


\bibitem[Vaswani et~al\mbox{.}(2023)]%
        {vaswani2023attention}
\bibfield{author}{\bibinfo{person}{Ashish Vaswani}, \bibinfo{person}{Noam Shazeer}, \bibinfo{person}{Niki Parmar}, \bibinfo{person}{Jakob Uszkoreit}, \bibinfo{person}{Llion Jones}, \bibinfo{person}{Aidan~N. Gomez}, \bibinfo{person}{Lukasz Kaiser}, {and} \bibinfo{person}{Illia Polosukhin}.} \bibinfo{year}{2023}\natexlab{}.
\newblock \bibinfo{title}{Attention Is All You Need}.
\newblock
\newblock
\showeprint[arxiv]{1706.03762}


\bibitem[Wang et~al\mbox{.}(2024)]%
        {mmlu_pro}
\bibfield{author}{\bibinfo{person}{Yubo Wang}, \bibinfo{person}{Xueguang Ma}, \bibinfo{person}{Ge Zhang}, \bibinfo{person}{Yuansheng Ni}, \bibinfo{person}{Abhranil Chandra}, \bibinfo{person}{Shiguang Guo}, \bibinfo{person}{Weiming Ren}, \bibinfo{person}{Aaran Arulraj}, \bibinfo{person}{Xuan He}, \bibinfo{person}{Ziyan Jiang}, {et~al\mbox{.}}} \bibinfo{year}{2024}\natexlab{}.
\newblock \showarticletitle{Mmlu-pro: A more robust and challenging multi-task language understanding benchmark}.
\newblock \bibinfo{journal}{\emph{arXiv preprint arXiv:2406.01574}} (\bibinfo{year}{2024}).
\newblock


\bibitem[Xu and Ma(2024)]%
        {CountLetters1}
\bibfield{author}{\bibinfo{person}{Nan Xu} {and} \bibinfo{person}{Xuezhe Ma}.} \bibinfo{year}{2024}\natexlab{}.
\newblock \showarticletitle{LLM The Genius Paradox: A Linguistic and Math Expert's Struggle with Simple Word-based Counting Problems}.
\newblock \bibinfo{journal}{\emph{arXiv preprint arXiv:2410.14166}} (\bibinfo{year}{2024}).
\newblock


\bibitem[Yang et~al\mbox{.}(2024)]%
        {ChatGPT_survey}
\bibfield{author}{\bibinfo{person}{Jingfeng Yang}, \bibinfo{person}{Hongye Jin}, \bibinfo{person}{Ruixiang Tang}, \bibinfo{person}{Xiaotian Han}, \bibinfo{person}{Qizhang Feng}, \bibinfo{person}{Haoming Jiang}, \bibinfo{person}{Shaochen Zhong}, \bibinfo{person}{Bing Yin}, {and} \bibinfo{person}{Xia Hu}.} \bibinfo{year}{2024}\natexlab{}.
\newblock \showarticletitle{Harnessing the Power of LLMs in Practice: A Survey on ChatGPT and Beyond}.
\newblock \bibinfo{journal}{\emph{ACM Trans. Knowl. Discov. Data}} \bibinfo{volume}{18}, \bibinfo{number}{6}, Article \bibinfo{articleno}{160} (\bibinfo{date}{April} \bibinfo{year}{2024}), \bibinfo{numpages}{32}~pages.
\newblock
\showISSN{1556-4681}
\urldef\tempurl%
\url{https://doi.org/10.1145/3649506}
\showDOI{\tempurl}


\bibitem[Yehudai et~al\mbox{.}(2024)]%
        {CountLetters2}
\bibfield{author}{\bibinfo{person}{Gilad Yehudai}, \bibinfo{person}{Haim Kaplan}, \bibinfo{person}{Asma Ghandeharioun}, \bibinfo{person}{Mor Geva}, {and} \bibinfo{person}{Amir Globerson}.} \bibinfo{year}{2024}\natexlab{}.
\newblock \showarticletitle{When Can Transformers Count to n?}
\newblock \bibinfo{journal}{\emph{arXiv preprint arXiv:2407.15160}} (\bibinfo{year}{2024}).
\newblock


\bibitem[Yuan et~al\mbox{.}(2023)]%
        {LLMmath}
\bibfield{author}{\bibinfo{person}{Zheng Yuan}, \bibinfo{person}{Hongyi Yuan}, \bibinfo{person}{Chuanqi Tan}, \bibinfo{person}{Wei Wang}, {and} \bibinfo{person}{Songfang Huang}.} \bibinfo{year}{2023}\natexlab{}.
\newblock \showarticletitle{How well do large language models perform in arithmetic tasks?}
\newblock \bibinfo{journal}{\emph{arXiv preprint arXiv:2304.02015}} (\bibinfo{year}{2023}).
\newblock


\bibitem[Zellers et~al\mbox{.}(2019)]%
        {CommonSensemeasuring}
\bibfield{author}{\bibinfo{person}{Rowan Zellers}, \bibinfo{person}{Ari Holtzman}, \bibinfo{person}{Yonatan Bisk}, \bibinfo{person}{Ali Farhadi}, {and} \bibinfo{person}{Yejin Choi}.} \bibinfo{year}{2019}\natexlab{}.
\newblock \showarticletitle{HellaSwag: Can a Machine Really Finish Your Sentence?}. In \bibinfo{booktitle}{\emph{Annual Meeting of the Association for Computational Linguistics}}.
\newblock


\end{thebibliography}

\clearpage

\appendix

\section{Appendix: Comparison real counts and detected counts of LLMs}

\begin{figure}[H]
    \centering
    \includegraphics[width=1\linewidth]{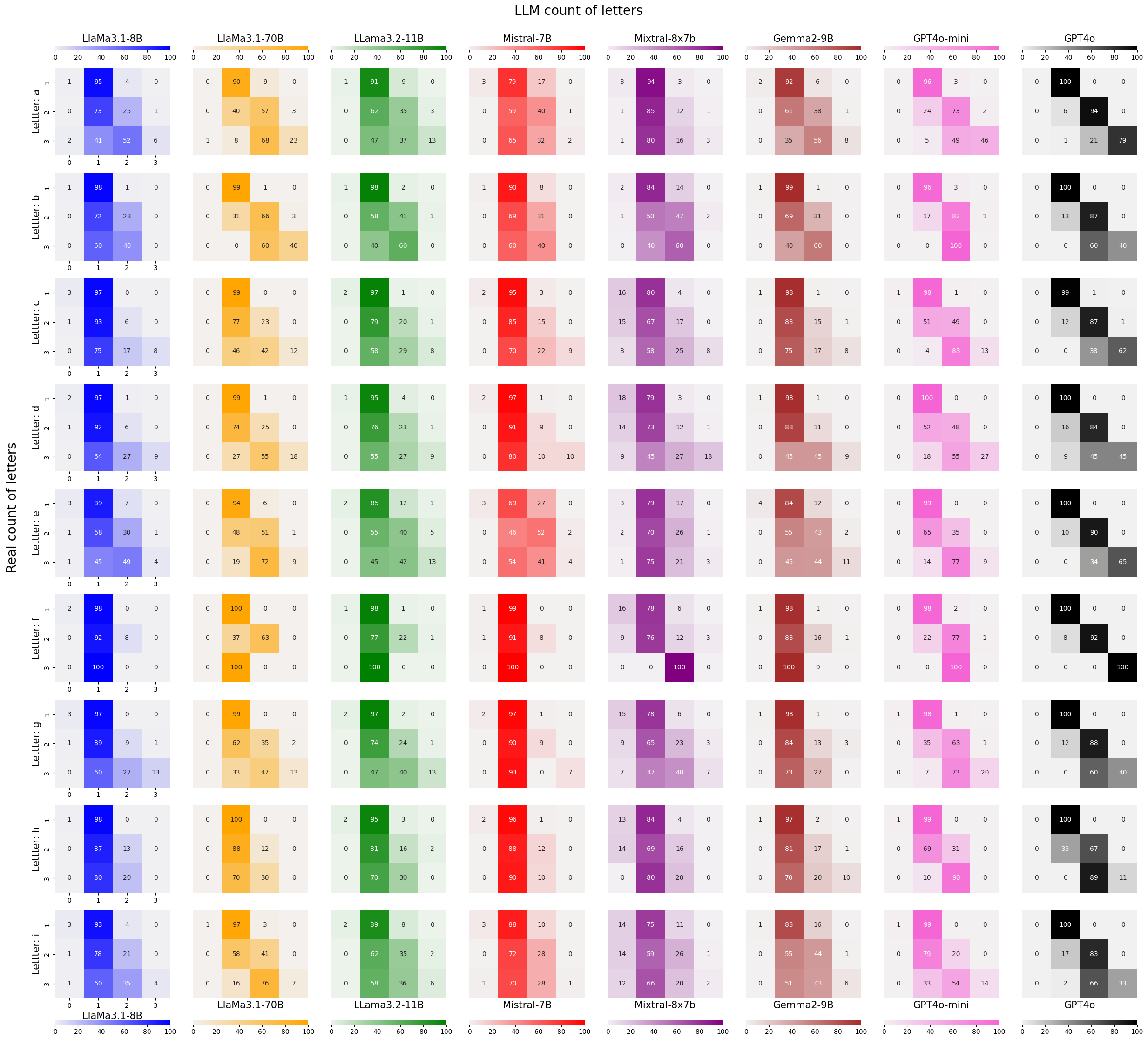}
    \caption{Heat matrix showing the percentage of actual letter occurrences (rows) vs. LLM-Detected Counts (columns). Letters a-i.}
    \label{fig:0to9_raw}
\end{figure}

\begin{figure}[H]
    \centering
    \includegraphics[width=1\linewidth]{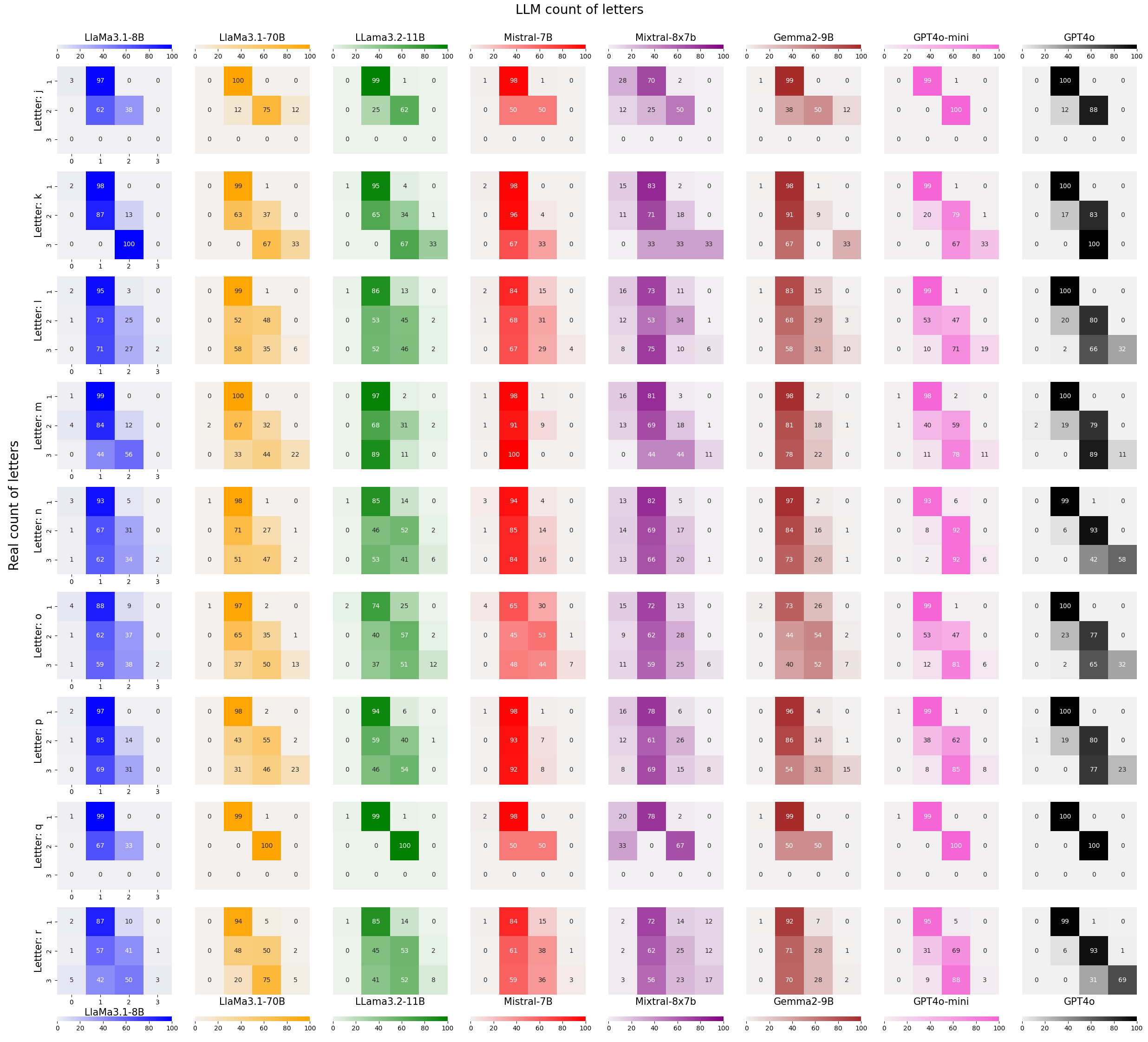}
    \caption{Heat matrix showing the percentage of actual letter occurrences (rows) vs. LLM-Detected Counts (columns). Letters j-r.}
    \label{fig:9to18_raw}
\end{figure}

\begin{figure}[H]
    \centering
    \includegraphics[width=1\linewidth]{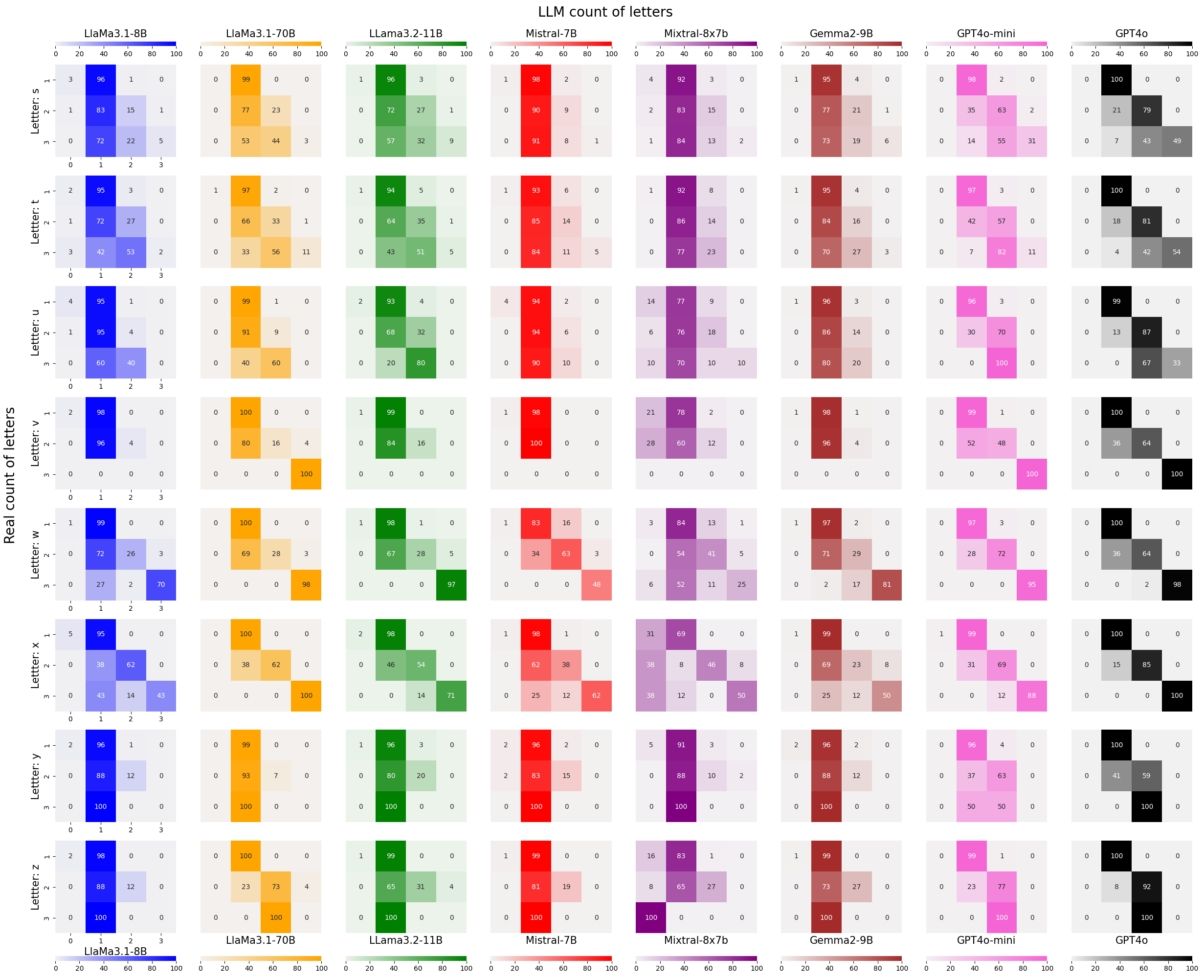}
    \caption{Heat matrix showing the percentage of actual letter occurrences (rows) vs. LLM-Detected Counts (columns). Letters s-z.}
    \label{fig:18to26_raw}
\end{figure}

Figures \ref{fig:0to9_raw}, \ref{fig:9to18_raw}, \ref{fig:18to26_raw} present the results separated by letters and models. For each letter and model, the rows represent the actual letter multiplicity (1,2,3) and the columns the count estimate of the model (0,1,2,3). The absence of errors corresponds to a descending staircase starting in the second column. A vertical line in the second column corresponds to a model that produces a count of one regardless of the actual letter multiplicity. It can be observed that the best performing models tend to produce maps that are closer to the ideal result. Most models have most errors by estimating a lower multiplicity than the actual value. The results vary for each letter, and some letters like x and w have better results for most models.   

\end{document}